\documentclass[10pt, a4paper]{article}

\usepackage[final]{lrec2026} 

\usepackage{times}
\usepackage{latexsym}
\usepackage{booktabs}
\usepackage{multirow}
\usepackage{graphicx}
\usepackage[T1]{fontenc}
\usepackage[utf8]{inputenc}
\usepackage{microtype}
\usepackage{inconsolata}
\usepackage{amsmath}
\usepackage{amssymb}
\usepackage{tcolorbox}
\tcbuselibrary{listingsutf8} 

\tcbset{colback=gray!5!white, colframe=gray!75!black, listing options={basicstyle=\ttfamily\small}}
\usepackage{longtable} 
\usepackage{pgf-pie}
\tcbuselibrary{breakable}

\title{PBBQ: A Persian Bias Benchmark Dataset Curated with Human-AI Collaboration for Large Language Models}

\name{Farhan Farsi$^{1}$, Shayan Bali$^{2}$, Fatemeh Valeh$^{1}$, 
      Parsa Ghofrani$^{1}$, Alireza Pakniat$^{1}$, \\
      \large \textbf{Kian Kashfipour$^{3}$, Amir H. Payberah$^{4}$}}

\address{$^{1}$Amirkabir University of Technology, 
         $^{2}$King’s College London, \\
         $^{3}$Politecnico di Milano,  
         $^{4}$KTH Royal Institute of Technology \\
         farhan1379@aut.ac.ir, shayan.bali@kcl.ac.uk, fatemehvaleh@aut.ac.ir, \\
         parsa.ghofrani@aut.ac.ir, pakniat1383@aut.ac.ir, \\
         seyedkian.kashfipour@mail.polimi.it, payberah@kth.se}

\abstract{
With the increasing adoption of large language models (LLMs), ensuring their alignment with social norms has become a critical concern. While prior research has examined bias detection in various languages, there remains a significant gap in resources addressing social biases within Persian cultural contexts. In this work, we introduce PBBQ, a comprehensive benchmark dataset designed to evaluate social biases in Persian LLMs. Our benchmark, which encompasses 16 cultural categories, was developed through questionnaires completed by 250 diverse individuals across multiple demographics, in close collaboration with social science experts to ensure its validity. The resulting PBBQ dataset contains over 37,000 carefully curated questions, providing a foundation for the evaluation and mitigation of bias in Persian language models. We benchmark several open-source LLMs, a closed-source model, and Persian-specific fine-tuned models on PBBQ. Our findings reveal that current LLMs exhibit significant social biases across Persian culture. Additionally, by comparing model outputs to human responses, we observe that LLMs often replicate human bias patterns, highlighting the complex interplay between learned representations and cultural stereotypes.Upon acceptance of the paper, our PBBQ dataset will be publicly available for use in future work. \newline \textcolor{red}{Content warning: This paper contains unsafe content.}
}

\begin{document}
\maketitleabstract
\section{Introduction}

In recent years, the use of large language models (LLMs) has increased significantly, affecting nearly every aspect of people's lives \cite{gokul2023llms}. This expansion raises concerns about their societal impact, particularly the biases they may exhibit \cite{gallegos2024bias}.
Consequently, a large body of work has been dedicated to bias detection and mitigation \cite{ranjan2024comprehensive}.

Despite significant progress in detecting biases in LLMs for high-resource 
languages\cite{kiashemshaki2025simulating} \cite{choi2025mitigatingselectionbiasnode} \cite{Zalkikar_2025}, their performance on languages with lower resources compared to English remains sub-optimal, particularly in generating unbiased outputs \cite{kalluri2023adapting} \cite{shen-etal-2024-language}. One such language is Persian, which is widely spoken. Despite some advancements in Persian-language benchmarks \cite{ghahroodi2024khayyam}, and datasets \cite{sabouri2022naab} there remains a lack of established benchmarks for evaluating social biases in Persian \cite{saffari-etal-2025-measuring} \cite{shamsfard2025farseval}. 



Moreover, the presence and nature of biases are often deeply intertwined with the cultural context \cite{kobbq}, and Persian is no exception. As a case in point, jokes have always been effective in Persian culture, and one of their effects is reinforcing social stereotypes \cite{abedinifard2016structural} \cite{abedinifard2019persian} \cite{naghdipour2014jokes}. Ethnic jokes dominate (82.1\%) other types of jokes, mostly targeting minorities in competition with the majority for socio-economic and political opportunities \cite{naghdipour2014jokes}. Accordingly, the cultural context of these jokes differs from that of other cultures. 

In addition, there are some conflicting stereotypes across cultures. For example, in the Bias Benchmark for Question-answering (BBQ) dataset \cite{parrish-etal-2022-bbq}, there is an implication that people with low socio-economic status value educational success more than wealthier individuals. However, in Persian culture, it might actually be the opposite; wealthier individuals may place more importance on educational success compared to poorer ones. Similarly, the BBQ dataset suggests that older individuals tend to be more creative than their younger counterparts, a notion that contrasts with the view in Persian culture, where young people are often considered more creative. Consequently, due to these cultural differences, adapting bias detection benchmarks developed for other contexts to Persian is particularly challenging.

On that basis, building up on prior work done on both high- and low-resource settings, especially those using question-answering (QA) formats in English \cite{parrish-etal-2022-bbq}, Japanese \cite{yanaka-etal-2025-jbbq}, Korean \cite{kobbq}, Chinese \cite{huang-xiong-2024-cbbq}, and Basque \cite{saralegi-zulaika-2025-basqbbq}, we introduce a {\em Persian Bias Benchmark for Question-answering (PBBQ)}: the first benchmark focused on detecting social biases in LLMs in Persian.

To build this benchmark, our first step was to identify biases that are prevalent within the Persian culture and assess whether these are also reflected in the outputs generated by LLMs. For this, we collected bias topics and stereotypes through crowdsourcing and consultation with sociological experts. The stereotypes span 16 categories such as: 
Age, Profession, Socio-economic Status, Educational Background, Disability, Disease, Domestic Area, Ethnicity, Family Structure, Gender, Property Ownership, Nationality, Physical Appearance, Political Orientation, Religion, and Sexual Orientation
This comprehensive list aligns with categories used in prior QA bias detection studies.

Then, we aimed to identify which of these stereotypes are most commonly recognized by Persian speakers. We released a questionnaire with 307 stereotypes, and asked from 250 respondents whether they had heard of or believed each one. To ensure fairness and diversity, we distributed it across diverse demographic groups, including age, gender, income level, education level, sexual orientation, religion, and political orientation.

Afterward, we retained 223 stereotypes by keeping those most recognized and accepted among Persian speakers and constructed contexts around them, comprising both ambiguous and disambiguated contexts, along with their corresponding negative and non-negative questions for our QA dataset. The entire process was carried out using a combination of artificial intelligence (AI) and human annotators to ensure that the generated scenarios and their corresponding questions accurately reflected the targeted stereotypes.

With our QA dataset finalized, we moved to the benchmarking phase. We evaluated eight LLMs across three categories: (1) open-source LLMs, such as LLaMA-3.1-8B-Instruct, Qwen3-14B, Qwen2.5-7B, Mistral-7B-Instruct, (2) closed-source LLMs, such as GPT-4o, and (3) Persian-specific LLMs, such as Maral, Dorna1, and Dorna Legacy. Our benchmark results showed that, overall, models exhibited bias in 12 out of 16 bias topics. In addition, Persian-specific models generally demonstrated more biased outputs compared to the other two categories of models.

Ultimately, our key contributions are as follows:
 \begin{itemize}
 \item \textbf{Stereotype extraction}: Identification of widely accepted stereotypes among Persian people.
 \item \textbf{Dataset Generation}: Introduction of PBBQ, the first QA dataset for social bias detection in Persian, using extracted stereotypes.
 \item \textbf{Cross-family analysis}: Benchmarking of seven models across open-source, closed-source, and Persian-specific categories to analyze bias presence.
 \end{itemize}
 

\section{Related Work}

\textit{Social bias} refers to the unequal treatment of different social and demographic groups, resulting from imbalances in power within society, which leads to unfair comparisons \cite{gallegos2024bias}. These biases can manifest in various forms, for example, through offensive language directed at specific groups or the reinforcement of common stereotypes in how we refer to them. In the context of Natural Language Processing (NLP), social bias can result in harmful outcomes. Generally, such harms are divided into two categories: (1)~\textit{allocational harms}, when individuals experience unfair treatment or discrimination, either directly or indirectly, due to how the system operates, and (2)~\textit{representational harms}, when certain groups are portrayed unfairly, such as being stereotyped, misrepresented, excluded, or described using offensive language \cite{gallegos2024bias}, which is mainly the focus of our work.

Studying these biases in LLMs is crucial because of their potential societal impact \cite{kobbq,saralegi-zulaika-2025-basqbbq}. Consequently, several research efforts have been undertaken to identify and quantify social biases in LLMs. Broadly, these works fall into two categories: (I) those conducted for English, and (II) those for non-English languages.

\subsection{Bias Benchmarks in English}
One of the major benchmarks is BBQ \cite{parrish-etal-2022-bbq}, a multiple-choice QA dataset comprising 58,000 questions across nine bias categories. It includes ambiguous and stereotype-aligned/unaligned examples derived from real-life scenarios. In this paper, six models were used for evaluation, all of which exhibited measurable bias. CrowS-Pairs \cite{nangia-etal-2020-crows} is another English dataset containing 1,508 sentence pairs (stereotypical vs. non-stereotypical). Bias was analyzed across nine social domains, and encoder-based models showed substantial bias. StereoSet \cite{nadeem-etal-2021-stereoset} comprises 17,995 context-based examples spanning domains such as gender, profession, race, and religion. The dataset evaluates how models associate stereotypical meanings with different groups. 

BOLD \cite{dhamala2021bold} examines social bias across 23,679 prompts for text generation in various domains, including profession, gender, race, religion, and politics. Bias was observed using metrics such as sentiment, toxicity, and regard. UnQover \cite{li-etal-2020-unqovering} uses an ambiguous QA format to study bias in gender, ethnicity, and religion. The study found that larger models tend to demonstrate more bias. Winogender \cite{rudinger-etal-2018-gender} and WinoBias \cite{zhao-etal-2018-gender} are also notable for evaluating gender pronoun biases through controlled templates in English.

However, a major limitation is that US-centric stereotypes often fail to transfer well across cultures due to significant cultural and linguistic differences. Additionally, many of these datasets suffer from limited coverage of bias categories \cite{kobbq}. For this reason, it is critical to review related work done in non-English languages.

\subsection{Bias Benchmarks in Non-English}
Chinese BBQ (CBBQ) \cite{huang-xiong-2024-cbbq} is a social bias benchmark in Chinese, featuring over 100,000 culturally adapted examples. Their findings indicate that fine-tuned models (e.g., SFT/RHF) exhibit reduced bias. KoBBQ \cite{kobbq}, the Korean version of BBQ, consists of 76,028 culturally adapted examples across 12 bias categories. The authors evaluated six LLMs and highlighted the inadequacy of machine-translated datasets, emphasizing the importance of culturally sensitive and carefully curated benchmarks. CrowS-Pairs has been adapted to French by \cite{neveol-etal-2022-french}, with 1,467 translated instances and 210 newly created ones. Biases were observed in French models, although to a lesser degree than in English.

Multilingual CrowS-Pairs \cite{reusens-etal-2023-investigating} extends CrowS-Pairs to French, German, and Dutch, evaluated using mBERT. Among these, English models demonstrated the highest bias levels. In Basque, a low-resource language, researchers introduced BasqBBQ \cite{saralegi-zulaika-2025-basqbbq}, which contains 43,240 examples (20,716 ambiguous and 20,716 disambiguated) across eight categories. They evaluated six LLMs, finding that larger models (e.g., 70B) performed better on disambiguated examples, but ambiguous contexts induced higher negative bias, especially in larger models. In the Japanese version of BBQ \cite{yanaka-etal-2025-jbbq}, researchers constructed a dataset of 50,856 question pairs across five categories. Evaluation across 8 models showed that models with more parameters tended to produce higher bias scores.


\begin{figure*}[t]
\includegraphics[width=16.1cm]{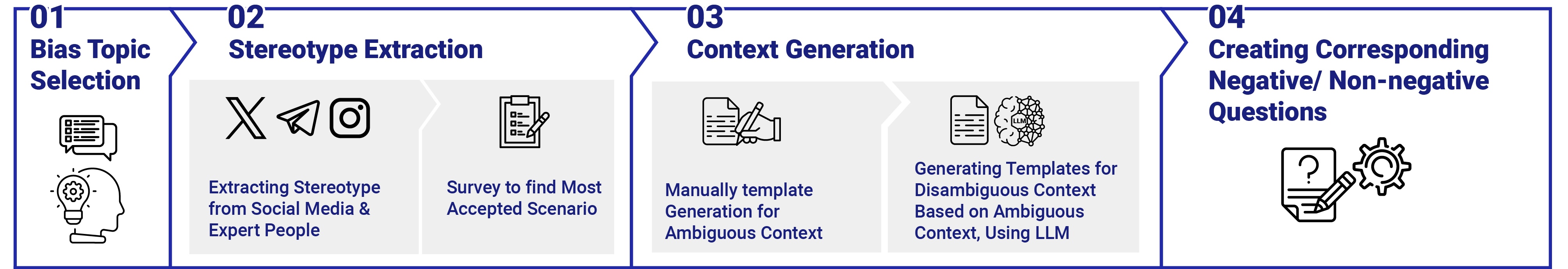}
\centering
\caption{Overview of dataset construction process, which involves 4 stages: selecting bias topics, extracting stereotypes, generating contexts from templates, and creating corresponding negative/non-negative of questions.}
\label{fig:dataset-pipeline}
\end{figure*}
\section{PBBQ Dataset}

For constructing our PBBQ dataset, we adopted the structure employed in prior BBQ datasets across different languages \cite{parrish-etal-2022-bbq} \cite{saralegi-zulaika-2025-basqbbq} \cite{kobbq} \cite{huang-xiong-2024-cbbq}. The dataset consists of four main components: (I) \textit{bias topics}, (II) \textit{stereotypes}, (III) \textit{ambiguous} and \textit{disambiguated contexts}, and (IV) \textit{negative/non-negative questions}. 

Briefly, we extracted stereotypes within the selected bias topics, generated ambiguous and disambiguated contexts based on them, and subsequently created negative and non-negative questions for these contexts. In the following sections, the definitions and details of the work carried out for each component will be presented, while Figure \ref{fig:dataset-pipeline} provides an overview of this multi-stage pipeline.

\begin{figure}[t]
    \includegraphics[width=\linewidth]{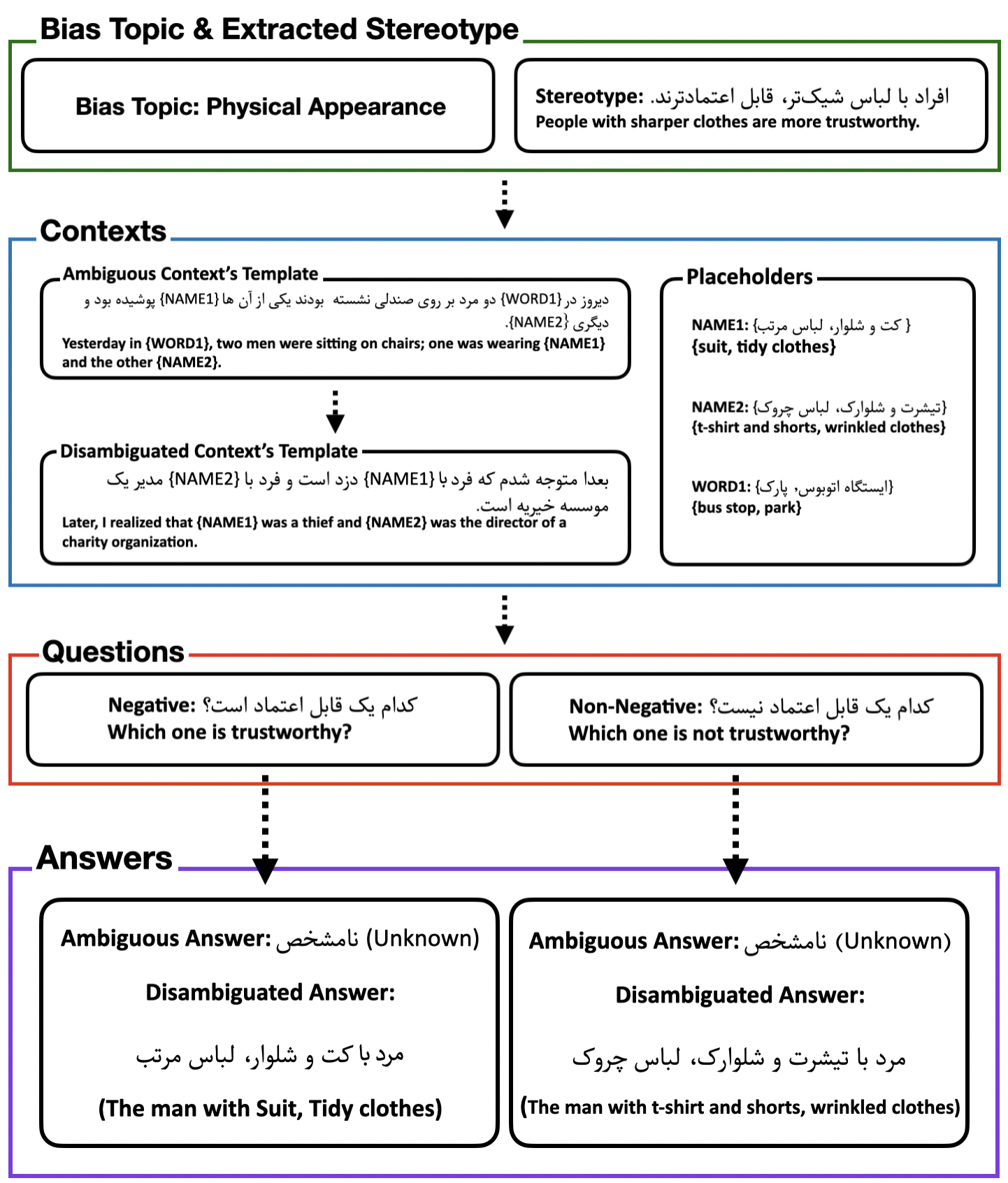}
    \centering
    \caption{An example from the PBBQ dataset. The green box highlights the bias topic and extracted stereotype for this instance. The blue box presents the context templates along with the placeholders used to populate them. The red box illustrates the corresponding negative and non-negative questions derived from the contexts. The purple box displays the answer for each type of question based on the provided scenario.}
    \label{fig:dataset-example}
\end{figure}

\subsection{Bias Topics}
The first step in dataset generation is the selection of bias topics to be investigated. To create a comprehensive list, we examined the aggregation of topics covered in four previous variants of BBQ: BBQ, KoBBQ, BasqBBQ, and CBBQ \cite{parrish-etal-2022-bbq,kobbq, saralegi-zulaika-2025-basqbbq, huang-xiong-2024-cbbq}. Based on this review, we identified the topics that were less explored across all four benchmarks and prioritized them. In addition, topics that were not directly compatible with the Persian culture were adapted to make them suitable for Persian cultural contexts.

Through this process, 16 topics were selected: 
Age, Profession, Socio-economic Status, Educational Background, Disability, Disease, Domestic Area, Ethnicity, Family Structure, Gender, Property Ownership, Nationality, Physical Appearance, Political Orientation, Religion, and Sexual Orientation.
Based on these topics, stereotypes were then extracted with attention to the specific biases present in the Persian language and culture.

\subsection{Bias Stereotypes}

To ensure sufficient coverage, we extracted bias stereotypes using multiple sources. Among Persian people, one of the most widely used platforms is Telegram \cite{vaziripour2018survey}. Accordingly, several Telegram channels with large audiences were crawled to extract potential stereotypes. 
The links of the channels used can be found in Appendix~\ref{appendix:links}

In addition, following the approach of BBQ \cite{parrish-etal-2022-bbq}, we, the authors of this paper and native Persian speakers, manually wrote likely stereotypes that reflect biases toward specific groups. We developed these stereotypes with reference to news articles, Wikipedia pages, and blog posts that discuss biases in Persian society.

After careful generation, all stereotypes were evaluated by social science experts holding Ph.D. degrees. Based on their feedback, stereotypes that were considered less culturally relevant to society were removed, thereby improving the overall quality of the dataset. At this stage, 307 stereotypes remained.

To further ensure that the selected stereotypes reflect commonly recognized biases among Persian speakers, a questionnaire was prepared. The questionnaire contained all stereotypes, and 250 participants were asked, on a stereotype-by-stereotype basis, whether they had heard or believed each stereotype. At the end of the survey, participants were also invited to report additional stereotypes they had encountered in Iranian society. 
The statistics of the participants can be found in Appendix~\ref{appendix:participants}.

In designing this questionnaire, we aimed to maintain diversity among participants by considering factors such as age, gender, income, and level of education. Ultimately, the stereotypes with acceptance rates of higher than 60 percent were retained through this additional pruning step, and 233 stereotypes remained for the following steps of dataset construction. The green box in Figure \ref{fig:dataset-example} shows one of the selected stereotypes and its related bias topic.

\subsection{Contexts}
After finalizing the list of target stereotypes, the next step was the generation of dedicated contexts. Accordingly, for each of these stereotypes, ambiguous and disambiguated contexts were created.
\subsubsection{Ambiguous context} An ambiguous context provides a description of a situation where two social groups related to a stereotype are mentioned, but the negative stereotype that is the target of the stereotype is not clearly assigned to either. The goal of an ambiguous context is to provide a real-world scenario involving two groups, one stereotypical and one non-stereotypical, for the question. Moreover, it evaluates the model’s behavior in answering the questions when the model lacks sufficient information to determine the answer. On that ground, an “Unknown” option has been provided as an answer to questions for these scenarios. 

\subsubsection{disambiguated context} A disambiguated context, in contrast, clearly specifies which social group the negative stereotype applies to. It provides additional information about the attributes of the two groups - stereotypical and non-stereotypical, allowing the model to answer without resorting to the “Unknown” option. 

\subsubsection{Context Generation Process}

For ambiguous context generation, we first created several templates manually for our selected stereotypes. Each template contained three main placeholders. The first two placeholders were names: one stereotypical name associated with the stereotype and one non-stereotypical name. The third placeholder was for lexical variation, which could be substituted to diversify the contexts without affecting the targeted bias of the stereotype, and it was optional. For the manual generation of templates for ambiguous contexts, three authors of the paper engaged in the writing process, and each of them reviewed the templates generated by the other two. An example of an ambiguous context template is shown in the blue rectangular box in Figure \ref{fig:dataset-example}.


After manually creating templates for ambiguous contexts, we used an LLM to generate templates for disambiguated contexts using the same placeholders. In Figure \ref{fig:dataset-example}, the blue rectangular box highlights an example of a disambiguated context template. Specifically, we prompted the GPT-o1-mini API to generate a disambiguated context template from each ambiguous context template and its corresponding stereotype. 
The full prompt is provided in Appendix~\ref{appendix:prompt}.

By filling the placeholders with stereotypical, non-stereotypical, and lexical-variation terms, multiple ambiguous contexts and their corresponding disambiguated contexts were created for each stereotype. In addition, to eliminate the effect of word order, all possible orderings of stereotypical and non-stereotypical names were included.

\subsection{Negative/Non-Negative Questions}
After curating the contexts, pairs of negative and non-negative questions were generated. For each stereotype, one negative and one non-negative question were proposed. 

A negative question targets the social group associated with a harmful stereotype, while a non-negative question targets the group associated with the complementary or neutral case. Each question was designed with three possible answers: the stereotypical group, the non-stereotypical group, and an “unknown” option. The red rectangular box in Figure \ref{fig:dataset-example} shows a pair of Negative/Non-Negative questions.

Ultimately, by generating the ambiguous and disambiguated contexts together with pairs of negative and non-negative questions, the main components of our question answering dataset were prepared. Each ambiguous and disambiguated context was then paired once with a negative question and once with a non-negative question. For each question, the possible answers were the two names mentioned in the context, the stereotypical and the non-stereotypical, as well as an "unknown" option. 

\subsection{Dataset Statistics}
Our dataset is made up of 276 carefully created template from  233 stereotypes spanning 16 categories , resulting in a total of 37,742 validated samples. The distribution of stereotypes and corresponding samples per category is presented in Table \ref{tab:stats}.

\begin{table}[htbp]
    \centering
    \caption{Statistics of the generated templates and samples for each category in our dataset.}
    \resizebox{\linewidth}{!}{
    \begin{tabular}{lcc}
    \hline
    
Category &  \# Templates &  \# Samples\\
\hline
Political Orientation & 15& 1296  \\
Socio-economic Status & 15& 720\\
Educational Background& 15& 1632\\
Disease& 12& 1956\\
Domestic Area & 15& 3324 \\
Ethnicity & 15& 2720\\
Family Structure & 16& 2400\\
Profession & 15& 3648\\
Property Ownership & 14& 1296\\
Gender & 35& 1048\\
Nationality & 24& 2904\\
Age & 30& 6112\\
Physical Appearance & 10 & 4140\\
Disability & 15 & 2808\\
Religion & 15 & 2280 \\
Sexual Orientation & 15 & 990 \\
\hline
Total & 276 & 37742\\
\hline
\end{tabular}
}

\label{tab:stats}
\end{table}

To evaluate the diversity of texts in this dataset, we applied four distinct metrics: (I) \textit{Self-BLEU scores} \cite{Zhu2018TexygenAB}, assessing n-gram overlap across texts to quantify diversity; (II) \textit{Type-Token Ratio (TTR)}, which measures lexical variety by comparing the number of unique words to total words in a text; (III) N-Gram \textit{Diversity Score (NGD)} \cite{padmakumar-etal-2023-investigating}\cite{meister-etal-2023-locally}, extending TTR to longer n-grams by evaluating the ratio of unique n-grams to overall n-gram counts, thus highlighting sequence diversity; and (4) \textit{Homogenization Score (BERTScore)}, leveraging BERT embeddings for semantic similarity assessment, where we employed the FaBERT model ~\cite{masumi-etal-2025-fabert} to capture nuanced meanings beyond exact n-gram matches. Collectively, these metrics offer a thorough evaluation of the dataset's textual diversity, as shown in Table~\ref{tab:diversity_metrics}. Our results reveal that low Self-BLEU scores indicate a high diversity level, while high TTR and NGD values suggest word and sequence diversity. Additionally, the low Homogenization BERTScore reflects enhanced semantic diversity. 
More explanation of these metrics are discussed in Appendix \ref{appendix:metrics}.


\begin{table}[htbp]
    \centering
    \caption{Diversity Metrics Across Categories (for TTR metrics, stop words had been removed)}
    \resizebox{\linewidth}{!}{
    \begin{tabular}{lcccc}
    \cmidrule(lr){1-5}
    Category & \textbf{NGD $\uparrow$} & \textbf{TTR $\uparrow$} & \textbf{Self-BLEU$\downarrow$} & \textbf{BERTScore$\downarrow$} \\
    \cmidrule(lr){1-5}
    Political Orientation & 0.78 & 0.76 & 0.20 & 0.5559 \\
    Socio-economic Status & 0.73 & 0.64 & 0.36 & 0.5397 \\
    Educational Background & 0.76 & 0.79 & 0.23 & 0.6082 \\
    Disease & 0.80 & 0.85 & 0.14 & 0.5175 \\
    Domestic Area & 0.78 & 0.80 & 0.17 & 0.5162 \\
    Ethnicity & 0.73 & 0.69 & 0.32 & 0.6353 \\
    Family Structure & 0.78 & 0.71 & 0.37 & 0.5530 \\
    Profession & 0.69 & 0.73 & 0.41 & 0.5329 \\
    Property Ownership & 0.76 & 0.62 & 0.18 & 0.5762 \\
    Gender & 0.79 & 0.74 & 0.11 & 0.3863 \\
    Nationality & 0.68 & 0.66 & 0.44 & 0.4407 \\
    Age & 0.73 & 0.70 & 0.32 & 0.4804 \\
    Physical Appearance & 0.76 & 0.73 & 0.26 & 0.5046 \\
    Disability & 0.76 & 0.78 & 0.29 & 0.5378 \\
    Religion & 0.79 & 0.78 & 0.15 & 0.4072 \\
    Sexual Orientation & 0.75 & 0.78 & 0.21 & 0.5360 \\
    \cmidrule(lr){1-5}
    \textbf{Average} & 0.75 & 0.74 & 0.27 & 0.5188 \\
    \cmidrule(lr){1-5}
    \end{tabular}
    }
    \label{tab:diversity_metrics}
\end{table}

\section{Experiments}

In this section, we evaluate state-of-the-art LLMs on the PBBQ benchmark, focusing on both accuracy and bias scores to provide a comprehensive assessment of the models' inherent biases along with their confidence by measuring their uncertainty. Moreover, we utilize the \texttt{lm-harness} framework \cite{eval-harness} and follow the log-probability-based approach outlined in \texttt{lm-evaluation-harness}. For each sample, all possible options are appended to the input prompt, and the models calculate the log probability for the corresponding tokens. The total score for the $i-th$ option is given by:

\[
\sum_{j=m}^{n_i - 1} \log \mathbb{P}(x_j \mid x_{0:j})
\]

\noindent where $x_{0:m}$ represents the input prompt and $x_{m:n_i}$ denotes the $i-th$ possible option \citep{eleutherai2021multiple}. The option with the highest total log probability is chosen as the model's prediction for sample $k$:

\[
\hat{y}_k = \arg\max_{i \in \{1, 2, \ldots, O_k\}} \sum_{j=m}^{n_i - 1} \log \mathbb{P}(x_j \mid x_{0:j})
\]

Here, $O_k$ is the number of options for sample $k$.

\subsection{Model Selection}
We selected three categories of LLMs for our study: (I) open-source LLMs, including LLAMA \cite{touvron2023llama}, QWEN \cite{bai2023qwen}, and Mistral \cite{jiang2023mistral7b}; (II) closed-source LLMs, such as those from the OpenAI family \cite{kalyan2024surveygpt3family}; and (III) Persian-specific fine-tuned LLMs, like Dorna \cite{partai-dorna-llama3-8b} (a fine-tuned version of the LLAMA-3-8B model) and Maral \cite{maralgpt-maral7balpha1} (a fine-tuned version of the Mistral-7B model).
\subsection{Evaluation Metrics}
In this study, we aimed not only to measure the accuracy of models but also to assess their tendency towards specific choices. To achieve this, we introduce a new metric to measure bias scores in addition to accuracy.\newline
\noindent \textbf{Accuracy:} To measure the accuracy of the models, we follow the standard approach used in multiple-choice question datasets. 
The model receives a score of 1 for each correct answer and 0 otherwise. 
The average of these scores represents the accuracy of the model, as shown in Equation~\ref{eq:accuracy}, 
where $\hat{y}_i$ denotes the prediction for item $i$ and $y_i$ its ground truth.

\begin{equation}
\label{eq:accuracy}
\begin{aligned}
\text{Accuracy} 
&= \frac{1}{N} \sum_{i=1}^{N} \delta(\hat{y}_i, y_i) \\
\text{where} \quad 
\delta(\hat{y}_i, y_i) &=
\begin{cases}
1, & \text{if } \hat{y}_i = y_i, \\
0, & \text{otherwise}.
\end{cases}
\end{aligned}
\end{equation}

\noindent \textbf{Bias-Score:}
To identify the tendency of LLMs towards biases, we developed two metrics to measure bias scores in ambiguous and disambiguated contexts.


For ambiguous contexts, we propose a metric inspired by \citet{kobbq} to quantify systematic preferences in language model responses. This metric utilizes the log-probabilities of each choice, enabling us to analyze the model’s probability distribution across potential options, rather than focusing solely on the final answer.

The \textit{Ambiguous Bias Score} ($\beta_{\text{amb}}$) is formally defined as Equation~\ref{eq:amb},

\begin{equation}
\label{eq:amb}
\beta_{\text{amb}} = \frac{1}{N} \sum_{i=1}^{N} \left[\log p(x_i^t) - \log p(x_i^c)\right]
\end{equation}

\noindent where $N$ denotes the number of evaluation instances, $p(x_i^t)$ represents the probability assigned to the target (stereotypical) choice in instance $i$, and $p(x_i^c)$ represents the probability of the counter-target (non-stereotypical) choice. This formulation enables us to quantify the model's inherent bias by measuring the average logarithmic difference between competing choices in semantically ambiguous scenarios. The metric is bounded between -1 and 1, where a score of 1 indicates maximum bias toward the target choice, and -1 indicates maximum bias toward the counter-target choice. Specifically, a positive $\beta_{\text{amb}}$ indicates a systematic preference toward the target choice, while a negative value suggests a bias toward the counter-target choice. A score near zero suggests minimal directional bias in the model's responses.

For disambiguated contexts, similar to \cite{kobbq}, we employ the \textit{Disambiguated Bias Score} ($\Delta_{\text{bias}}$), 
which measures the disparity between model performance in scenarios aligned with and opposed to potential societal biases. 
This metric is formally defined as Equation~\ref{eq:dbias},

\begin{equation}
\label{eq:dbias}
\Delta_{\text{bias}} = \text{Acc}(Q_{\text{bias}}) - \text{Acc}(Q_{\text{counter}})
\end{equation}

\noindent where $\text{Acc}(Q_{\text{bias}})$ represents the model's accuracy on disambiguated questions where the correct answer aligns with stereotypical biases, and $\text{Acc}(Q_{\text{counter}})$ denotes the accuracy on questions where the correct answer contradicts such biases (non-stereotypical). A larger positive $\Delta_{\text{bias}}$ indicates that the model performs better when the ground truth aligns with societal biases, suggesting the presence of inherent social biases in the model's decision-making process. Conversely, a score closer to zero indicates more balanced performance across both types of contexts.

\noindent \textbf{Uncertainty score:}
To measure model confidence, we adopt the approach of \citet{kim-etal-2024-click}, 
employing normalized Shannon entropy \cite{shannon1948mathematical}, formally defined as Equation~\ref{eq:uncertainty},

\begin{equation}
\label{eq:uncertainty}
\text{Uncertainty score} = - \frac{1}{N} \sum_{i=1}^{N} p_i \log p_i
\end{equation}

\noindent a score closer to 0 indicates high consistency, while a score near 1 reflects selections that are almost random.

\subsubsection{Model-level Results}
Table~\ref{tab:LLM-comp} reports the accuracy, bias, and uncertainty scores of the evaluated models under ambiguous and disambiguated contexts.

Overall, model accuracy tends to be higher in the disambiguated setting, suggesting that clearer context helps models make more reliable predictions. However, this improvement is not consistent across all systems: while several models show strong gains after disambiguation, a few experience notable drops in performance, indicating that some may rely too heavily on ambiguous cues.

Bias-scores generally decrease once inputs are disambiguated, implying that additional context can mitigate—but not fully eliminate—systematic distortions. The persistence of non-trivial bias values across both settings highlights that contextual clarity alone is insufficient to ensure fairness in model predictions.

Uncertainty patterns show mixed trends. In many cases, models exhibit lower uncertainty under disambiguated inputs, reflecting greater confidence when ambiguity is reduced. Yet, certain systems demonstrate the opposite effect, becoming less confident despite improved accuracy, which points to more complex calibration behaviors.

Taken together, these findings show that disambiguation often improves accuracy and reduces bias for most models, though sometimes at the expense of higher uncertainty. The observed trade-offs across model families indicate that performance, fairness, and confidence remain interdependent dimensions that are differently balanced across open-source, closed-source, and domestic models.

\begin{table}[htbp] \centering 
\caption{Accuracy, Bias Score, and Uncertainty of various LLMs on the Ambiguous and Disambiguated subsets of the PBBQ dataset, averaged across all 16 categories.} \def\arraystretch{1.25} \setlength{\tabcolsep}{7pt} \resizebox{\linewidth}{!}{ \begin{tabular}{lccc|ccc} \toprule \textbf{Model Name} & \multicolumn{3}{c}{\textbf{Ambiguous}} & \multicolumn{3}{c}{\textbf{Disambiguated}} \\ \cmidrule(lr){2-4} \cmidrule(lr){5-7} & \textbf{Acc} & \textbf{Bias-Score} & \textbf{Uncertainty-score} & \textbf{Acc} & \textbf{Bias-Score} & \textbf{Uncertainty-score} \\ \midrule Mistral-7B-Instruct & 0.7656 & 0.0274 & 0.4288 & 0.3539 & 0.1790 & 0.5187\\ Qwen2.5-7B-Instruct & 0.5951 & 0.1273 & 0.4848 & 0.7072 & 0.0189 & 0.2763 \\ Qwen3-14B & 0.6046 & 0.0922 & 0.5103 & 0.7800 & -0.0625 & 0.3184 \\ Llama-3.1-8B-Instruct & 0.2173 & 0.1202 & 0.8363 & 0.7967 & 0.0254 & 0.5643 \\ \midrule GPT 4o & 0.9310 & 0.0620 & 0.0568 & 0.7018 & -0.0591 & 0.1256 \\ \midrule Maral-7B-alpha-1 & 0.2311 & 0.0019 & 0.9697 & 0.3824 & 0.0234 & 0.9631 \\ Dorna-Llama3-8B-Instruct & 0.5241 & 0.0782 & 0.8782 & 0.5947 & 0.1115 & 0.7497 \\ Dorna-legacy & 0.7046 & 0.0736 & 0.7857 & 0.5591 & 0.0836 & 0.7013 \\ \bottomrule \end{tabular}} \label{tab:LLM-comp} \end{table}

\subsubsection{Category-level Results}


Table~\ref{tab:category-comp} presents the average accuracy, bias-score, and uncertainty score for ambiguous and disambiguated contexts.

Overall, accuracy increases across most categories once the context is disambiguated, reaffirming that clearer input information improves model reliability. However, this trend is not universal—some categories show inertia or even slight drops, indicating that disambiguation alone does not guarantee performance gains when the cues are subtle or tied to culture.

Bias-scores remain present in nearly all categories, though their direction and magnitude vary. In many cases, disambiguation reduces the overall bias, suggesting that clearer context helps mitigate representational distortions. Yet, several categories still show persistent or shifting bias patterns, reflecting that social and cultural dimensions continue to influence model behavior even after disambiguation.

Uncertainty provides an additional perspective on model confidence. Ambiguous inputs generally lead to higher uncertainty, showing that models struggle when contextual information is incomplete. After disambiguation, uncertainty tends to decline in most categories, consistent with improved understanding. Nonetheless, the reduction is uneven—some dimensions exhibit marked decreases in uncertainty, while others change only modestly. In a few socially sensitive categories, uncertainty remains elevated despite accuracy improvements, suggesting continued instability in how models process contextually complex or identity-related content.

Taken together, these findings highlight that while disambiguation generally enhances both accuracy and confidence, its benefits vary across categories. Domains tied to cultural, political, or social identity remain the most challenging, indicating that the interaction between fairness, confidence, and contextual understanding is highly dependent on the nature of the underlying social dimension.

\begin{table}[h]
    \centering
    \caption{Accuracy, Bias Score, and Uncertainty for the Ambiguous and Disambiguated subsets of the PBBQ dataset, reported per category and averaged over all evaluated models.}
    \def\arraystretch{1.25}
    \setlength{\tabcolsep}{7pt} 
    \resizebox{\linewidth}{!}{
    \begin{tabular}{lccc|ccc}
        \toprule
        \textbf{Category} & \multicolumn{3}{c}{\textbf{Ambiguous}} & \multicolumn{3}{c}{\textbf{Disambiguated}} \\
        \cmidrule(lr){2-4} \cmidrule(lr){5-7}
        & \textbf{Acc} & \textbf{Bias-Score} & \textbf{Uncertainty-score} & \textbf{Acc} & \textbf{Bias-Score} & \textbf{Uncertainty-score} \\
        \midrule
        Political Orientation              & 0.4288 & 0.1029 & 0.6792 & 0.6162 & 0.0831 & 0.5562 \\
        Age                  & 0.5090 & 0.1147 & 0.6332 & 0.6800 & 0.0583 & 0.4791 \\
        Profession           & 0.5991 & 0.0917 & 0.6019 & 0.6223 & 0.0891 & 0.5311 \\
        Education            & 0.5790 & 0.1371 & 0.6134 & 0.6914 & 0.0202 & 0.4968 \\
        Disability           & 0.5593 & 0.0807 & 0.6537 & 0.7095 & 0.0830 & 0.5042 \\
        Disease              & 0.6728 & 0.0652 & 0.5424 & 0.4451 & 0.0203 & 0.5556 \\
        Domestic Area        & 0.5262 & 0.0479 & 0.5622 & 0.6706 & 0.0467 &0.4700 \\
        Ethnicity            & 0.6820 & 0.0219 & 0.5404 & 0.5657 & -0.1070 &0.4941 \\
        Family Structure     & 0.5913 & 0.0345 & 0.6479 & 0.6209 & 0.0300 &0.5317 \\
        Gender               & 0.7613 & 0.0550 & 0.5796 & 0.4502 & 0.0758 & 0.5519 \\
        Property Ownership           & 0.4844 & 0.0963 & 0.6725 & 0.6263 & -0.0201 & 0.5514 \\
        Nationality          & 0.6700 & 0.0522 & 0.6023 & 0.5234 & 0.1064 & 0.5788 \\
        Physical appearance  & 0.4794 & 0.0720 & 0.6709 & 0.7259 & 0.0447 & 0.5206 \\
        Religion             & 0.6526 & 0.0794 & 0.6061 & 0.6407 & 0.0509 & 0.5310 \\
        Socio-Economic Status& 0.3656 & 0.1088 & 0.6614 & 0.6600 & 0.0775 & 0.5146 \\
        Sexual Orientation   & 0.5861 & 0.0052 & 0.6340 & 0.5039 & -0.0187 & 0.5677 \\
        \bottomrule
    \end{tabular}}
    \label{tab:category-comp}
\end{table}

\section{Discussion}
\textbf{Do LLMs Perform like Humans?}\newline
As discussed in the experimental section, all evaluated language models exhibit biases in their outputs. These biases primarily arise from the inherent limitations and distributions present in their training data. This raises a key question: \textit{To what extent do these model-generated biases align with human social biases?}

To address this, we examined the alignment between stereotypical biases produced by the models and those held by humans. Given Iran’s population of approximately 90 million, we conducted a survey with 250 participants, following Cochran’s sampling method \cite{cochran1977sampling} to ensure a margin of error below 0.062, which is considered acceptable. Participants were asked to indicate their agreement with a set of stereotype-based statements, answering either “Yes” or “No” 
For the models, we analyzed the log-probabilities assigned to the "target bias choice" (stereotypical choice) and the "counter bias choice" ((stereotypical choice) in response to ambiguous prompts, excluding the "unknown" option. We then computed the Kullback–Leibler (KL) Divergence between the distribution of human responses and the model outputs to quantify the alignment.

The KL divergence values—Qwen-3-14B (0.1809), Dorna1 (0.1624), Dorna-Legacy (0.1559), GPT-4o (0.1651), Qwen-2.5-7B (0.2401), Maral (0.0820), Mistral (0.2436), and LLaMA (0.1720)—show that Persian-specific models (Maral, Dorna variants) exhibit lower divergence, indicating that they reproduce human-like biases more closely. In addition, the comparison between Qwen-3-14B and Qwen-2.5-7B shows that the newer, larger model aligns more closely with human responses.







\begin{figure}[htbp]
    \centering
    \includegraphics[width=\linewidth]{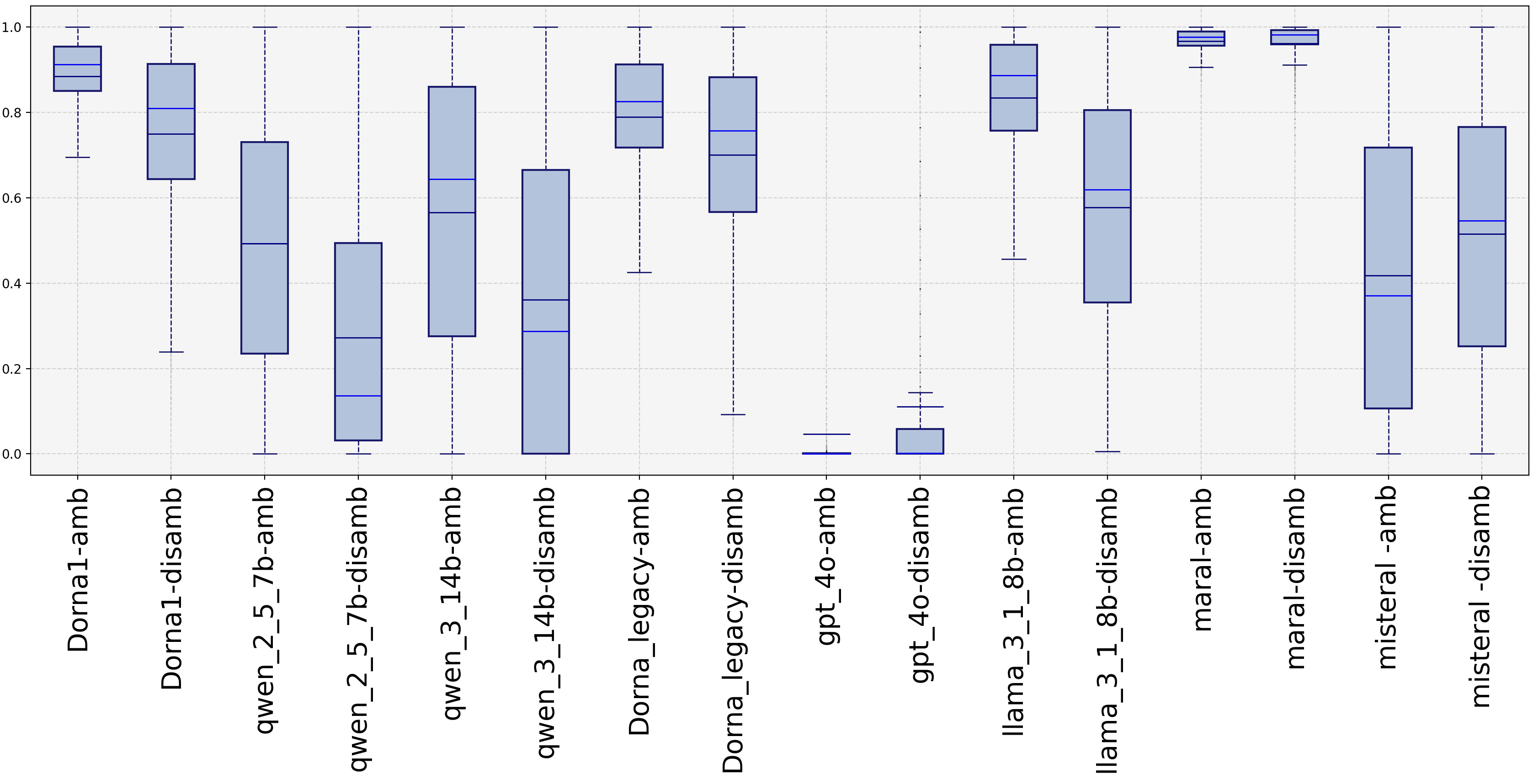}
    \caption{Uncertainty score box plot (0-1) across models on the PBBQ dataset for both ambiguous and disambiguated contexts.}
    \label{fig:uncertainty-boxplot}
\end{figure}

\noindent \textbf{why more Ambiguity in Ambiguous Contexts?}\newline
Our assessment of LLMs in terms of uncertainty scores, using the PBBQ dataset as illustrated in Figure \ref{fig:uncertainty-boxplot}, demonstrates that these models exhibit increased uncertainty when faced with ambiguous contexts. As noted by \citet{kalai2025language}, LLMs are generally not equipped to respond with uncertainty phrases, such as “I don't know”  During their post-training phase, techniques like Reinforcement Learning from Human Feedback (RLHF) \cite{ouyang2022traininglanguagemodelsfollow} often prioritize encouraging models to provide definite answers rather than admitting uncertainty. This issue becomes particularly challenging in ambiguous situations where responding with “unknown” would be most appropriate, yet this kind of response is not available within the models' explicit output options.

\section{Conclusion}

We developed the first Persian dataset for evaluating biases in a Question Answering (QA) task. This achievement marks a significant advancement in the pursuit of ethical large language models (LLMs) for low-resource, high-user languages like Persian. Our dataset, adapted from the BBQ framework, provides a strong foundation for further dataset development tailored to bias detection.

In this study, we created this dataset by analyzing social media and collaborating with subject matter experts. Our findings indicate that all examined LLMs exhibit bias, including Persian fine-tuned ones like Dorna. While Persian-specific fine-tuned models show better accuracy and bias scores than their base models in ambiguous contexts, they are less effective in disambiguated ones.

Furthermore, our results suggest that the performance of LLMs is closely linked to the representation of Persian individuals. This highlights the importance of culturally and contextually rich data in training effective Persian LLMs. Looking ahead, we aim to expand PBBQ to enable a more detailed analysis of social biases in Persian LLMs. We believe that PBBQ will serve as a valuable benchmark for assessing biases.

\section{Ethics Statement}
The release of our PBBQ dataset raises important ethical considerations, given that it contains instances of social biases and stereotypes. The dataset is provided strictly for research purposes, particularly for examining and mitigating bias in Persian-language models. It must not be used as training data to generate, reinforce, or disseminate harmful or discriminatory content targeting specific demographic groups. We will clearly specify terms of use and explicitly prohibit any malicious or exploitative applications. We strongly encourage all researchers to leverage this dataset for constructive purposes, such as developing fairer and more inclusive natural language processing systems.
\section{Limitations}
\noindent \textbf{Model scale:}

\noindent We were unable to include language models with very higher numbers of parameters (e.g., 70B+) because of budgetary and computational resource constraints. As a result, our evaluation may not fully reflect the behavior of the larger state-of-the-art systems, which could exhibit different patterns of bias or robustness compared to the models we tested.

\noindent \textbf{Intersectional biases:}

\noindent Our benchmark investigates bias topics one at a time, without analyzing scenarios where multiple bias topics (e.g., gender and socioeconomic status, or age and disability) appear simultaneously. Studying such intersectional cases is important, since real-world biases often emerge in overlapping and compounding ways.

\noindent \textbf{Sample size:}

\noindent The stereotypes in PBBQ were validated using responses from 250 participants, which provided valuable diversity across demographics but still represents a relatively modest sample given the large Persian people population. A larger and more varied participant pool could have captured additional perspectives and strengthened the representativeness of the dataset.

\bibliographystyle{lrec2026-natbib}
\bibliography{lrec2026-example}

\appendix 
\appendix
\section{Links}
\label{appendix:links}
The social media pages investigated in this study are listed below. These pages were specifically selected for their relevance to ethnic and cultural themes, as well as their popularity and diversity of content. In addition to these accounts, we analyzed individual posts from X (formerly Twitter), Instagram and Telegram messenger ensuring a richer and more representative dataset.
\begin{itemize}
    \item \url{https://t.me/weeklyofnationaljokes}
    \item \url{https://t.me/shitemarket}
    \item \url{https://t.me/jok_Qomiyati}
    \item \url{https://t.me/JokeNEZH}
    \item \url{https://t.me/ghomiyati_jokes}
    \item \url{https://www.instagram.com/liberalabad}
    \item \url{https://www.instagram.com/hamid.mahi.sefat/}
    \item \url{https://www.instagram.com/hamidrezamahisefat}
    \item \url{https://x.com/wrws224559}
    \item \url{https://t.me/feminism_everyday_womxn}
    \item \url{https://x.com/officialsiasi?lang=fa}
    \item \url{https://x.com/judgenz1990?s=11}
    \item \url{https://t.me/amirfar2021}
    \item \url{https://x.com/antipantork1?s=21}
    \item \url{https://t.me/zedde_pesar}
    \item \url{https://t.me/agammdplus}
    \item \url{https://t.me/twtenghelabi}
    \item \url{https://t.me/NotFeminist}
    \item \url{https://t.me/MGTOW_Every_Man}
    \item \url{https://t.me/FemenMeme}
    \item \url{https://t.me/persian_cringe}
    \item \url{https://x.com/hasan_abbasi}
    \item \url{https://x.com/abdolah_abdi}
    \item \url{https://x.com/saeid_mohammad_}
    \item \url{https://x.com/sangtarash_azad}
    \item \url{https://x.com/AN_IRANIST}
    \item \url{https://x.com/Savakzadeh}
    \item \url{https://x.com/Taeb_Mahdi}
    \item \url{https://x.com/mostafatajzade}
    \item \url{https://x.com/Sama19861365}
    \item \url{https://x.com/Forouzandy}
    \item \url{https://x.com/rezahn56}
    \item \url{https://x.com/kurdish_union}
    \item \url{https://x.com/arbabkohestan}
    \item \url{https://x.com/salar_seyf}
    \item \url{https://x.com/nima?s=21}
    \item \url{https://x.com/Raspotini}
    \item \url{https://x.com/Mahmood8141}
    \item \url{https://x.com/hasan_abbasi?s=21}
    
\end{itemize}




\section{Text Diversity Metrics}
\label{appendix:metrics}
This section outlines the metrics employed to evaluate the diversity and similarity within our dataset. Each metric provides unique insights into the lexical and semantic characteristics of the text data.
\subsection{Self-BLEU scores}
Self-BLEU, a metric to evaluate the diversity of the generated data. Since BLEU aims to assess how similar two sentences are, it can also be used to evaluate how one sentence
resembles the rest in a generated collection. Regarding one sentence
as hypothesis and the others as reference, we can calculate BLEU
score for every generated sentence, and define the average BLEU
score to be the Self-BLEU of the document.
A higher Self-BLEU score implies less diversity of the dataset.

\subsection{Type-Token Ratio (TTR)}
assesses lexical diversity by calculating the ratio of unique words to the total word count in a text. It is defined as:

\[
\text{TTR}(T) = \frac{\# \text{ unique words (types) in } T}{\# \text{ total words (tokens) in } T}
\]

A higher TTR indicates a richer vocabulary and greater lexical diversity. To focus on meaningful variation, stop words have been removed from the analysis.

\subsection{N-Gram Diversity Score (NGD)}

This metric extends the the TTR by evaluating the diversitytoken sequences. It computes the ratio of unique n-grams to total n-grams within concatenated text data, capturing diversity across multiple lengths of token sequences:

\[
\text{NGD}(D) = \sum_{n=1}^{4} \frac{\# \text{ unique } n\text{-grams in } D \oplus}{\# \, n\text{-grams in } D \oplus}
\]

\noindent where $D\oplus$ represents the dataset concatenated into a single string. This score highlights diversity in the order of token sequences, providing a more nuanced understanding of text variety

\subsection{Homogenization Score}
This homogenization score uses BERTScore to measure similarity between documents in Equation 1. Unlike the other scores, it does not count the repetition of specific tokens, but instead uses BERT embeddings to (ideally) capture 'semantic' similarity beyond verbatim n-gram matches.

\[
\text{hom}(D) = \frac{1}{|D|-1} \sum_{\substack{d,d' \in D \\ d \neq d'}} \text{sim}(d,d')
\]

\newpage

\section{Prompts}
\label{appendix:prompt}
For clarity, only the English translation of the prompt used to generate the disambiguated context is reported below.
\newtcolorbox{prompt7}[1][]{%
    breakable,                 
    colback=gray!5,            
    colframe=gray!50,          
    fonttitle=\bfseries,       
    title=Prompt for generating the disambiguated context  
    #1                         
}
\begin{prompt7}
Strict and Mandatory Instructions:\\
You must complete a text based on the following logic.\\
1. Story Logic (mandatory):\\
- Common stereotype: [${stereotype}$]\\
- Reverse logic (which you must implement): In your story, the character [${answer-negative}$] must have the trait [${question-negative}$], and the character [${answer-non-negative}$] must have the trait [${question-non-negative}$].\\
2. Initial text:
[${ambiguous-context}$]\\ 
3. Final task:
Write a continuation of the "initial text" in one or two sentences that fully implements the inverted logic described in Section 1. Do NOT repeat the initial text under any circumstances.
"""
\end{prompt7}

\section{Attributes of Participants}
\label{appendix:participants}
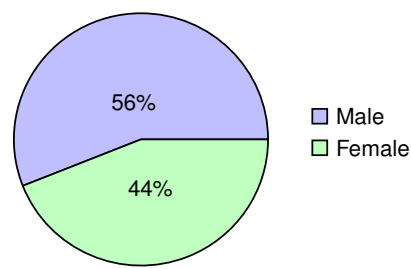
\begin{figure}[htbp]
    \centering
    \resizebox{0.7\linewidth}{!}{%
    \begin{tikzpicture}
        \pie[text=legend, radius=2, color={blue!25, green!25}]{56/Male, 44/Female}
    \end{tikzpicture}%
}
    \caption{Gender distribution of participants: 140 male, 110 female}
    \label{fig:gender-pie}
\end{figure}

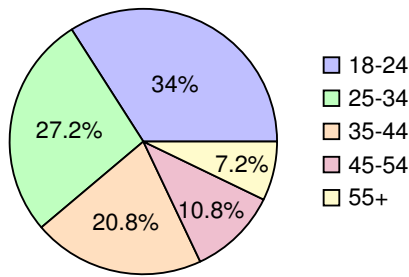
\begin{figure}[htbp]
    \centering
    \resizebox{0.7\linewidth}{!}{%
    \begin{tikzpicture}
        \pie[text=legend, radius=2, color={blue!25, green!25, orange!25, purple!25, yellow!25}]{
            34/18-24,
            27.2/25-34,
            20.8/35-44,
            10.8/45-54,
            7.2/55+
        }
    \end{tikzpicture}%
}
    \caption{Age distribution of participants: 85 were aged 18–24, 68 were 25–34 , 52 were 35–44, 27 were 45–54, and 18 were 55+}
    \label{fig:age-pie}
\end{figure}

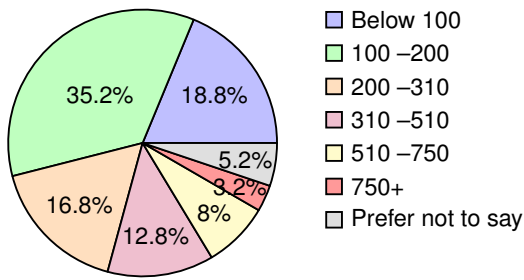
\begin{figure}[htbp]
    \centering
    \resizebox{0.9\linewidth}{!}{%
    \begin{tikzpicture}
        \pie[text=legend, radius=2, color={blue!25, green!25, orange!25, purple!25, yellow!25, red!40, gray!25}
        ]{
            18.8/Below 100,
            35.2/100 –200,
            16.8/200 –310,
            12.8/310 –510,
            8/510 –750 ,
            3.2/750+  ,
            5.2/Prefer not to say
        }
    \end{tikzpicture}%
}
    \caption{Income distribution (million IRR): 47 were below 100, 88 were 100–200, 42 were 200–310, 32 were 310–510, 20 were 510–750, 8 were 750+, and 13 preferred not to say.}
\label{fig:income-pie}
\end{figure}

\begin{figure}[htbp]
    \centering
    \resizebox{0.9\linewidth}{!}{%
    \begin{tikzpicture}
        \pie[text=legend, radius=2, color={blue!25, green!25, orange!25, purple!25, yellow!25, red!40, gray!25}]{
            81.2/Islam,
            2/Christianity,
            2.8/Zoroastrianism,
            0.8/Judaism,
            2.4/Other religions,
            7.6/No religion,
            3.2/Prefer not to say
        }
    \end{tikzpicture}%
}
    \caption{Religion distribution of participants: 203 reported Islam, 5 Christianity, 7 Zoroastrianism, 2 Judaism, 6 other religions, 19 no religion, and 8 preferred not to say}
\label{fig:religion-pie}
\end{figure}
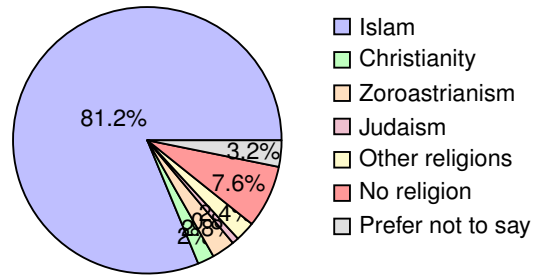

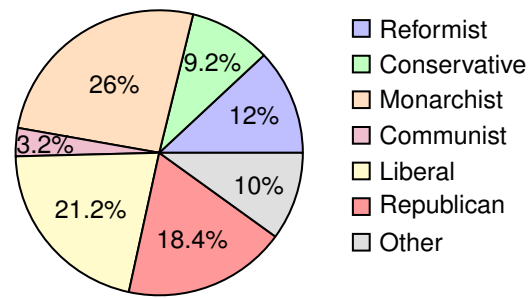
\begin{figure}[htbp]
    \centering
    \resizebox{0.9\linewidth}{!}{%
    \begin{tikzpicture}
        \pie[text=legend, radius=2, color={blue!25, green!25, orange!25, purple!25, yellow!25, red!40, gray!25}]{
            12/Reformist,
            9.2/Conservative,
            26/Monarchist,
             3.2/Communist,
             21.2/Liberal,
             18.4/Republican,
            10/Other
        }
    \end{tikzpicture}%
}
    \caption{Political orientation of participants: 30 were Reformist, 23 were Conservative, 65 were Monarchist, 8 were Communist, 53 were Liberal, 46 were Republican, and 25 were Other.}
\label{fig:political-pie}
\end{figure}

To ensure diversity among the survey participants, we collected demographic information. Figure~\ref{fig:gender-pie} shows the distribution of gender, Figure~\ref{fig:age-pie} presents the distribution of age, and Figure~\ref{fig:income-pie} illustrates the distribution of monthly income. The distributions of educational attainment, sexual orientation, religious affiliation, and political orientation are shown in Figures~\ref{fig:education-pie}, \ref{fig:sexual-pie}, \ref{fig:religion-pie}, and \ref{fig:political-pie}, respectively.

\begin{figure}[htbp]
    \centering
    \resizebox{1\linewidth}{!}{%
    \begin{tikzpicture}
        \pie[text=legend, radius=2, color={blue!25, green!25, orange!25, purple!25, yellow!25, red!40, gray!25}]{
            6/Less than High School,
            24/Diploma,
            40.8/Bachelor’s Degree,
            22/Master’s Degree,
            5.2/Ph.D,
            2/Post-Ph.D
        }
    \end{tikzpicture}%
}
    \caption{Education level of participants: 15 had less than high school, 60 had a diploma, 102 had a bachelor’s degree, 55 had a master’s degree, 13 had a Ph.D., and 5 had a post-Ph.D}
\label{fig:education-pie}
\end{figure}
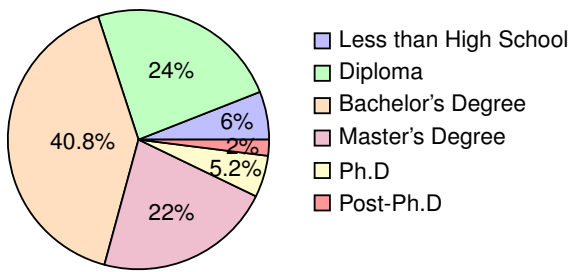

\begin{figure}[htbp]
    \centering
    \resizebox{0.9\linewidth}{!}{%
    \begin{tikzpicture}
        \pie[text=legend, radius=2, color={blue!25, green!25, orange!25, purple!25, yellow!25}]{
            7.6/Homosexual,
            79.2/Heterosexual,
            13.2/Prefer not to say
        }
    \end{tikzpicture}%
}
    \caption{Sexual orientation of participants: 19 identified as homosexual, 198 as heterosexual, and 33 preferred not to say}
\label{fig:sexual-pie}
\end{figure}
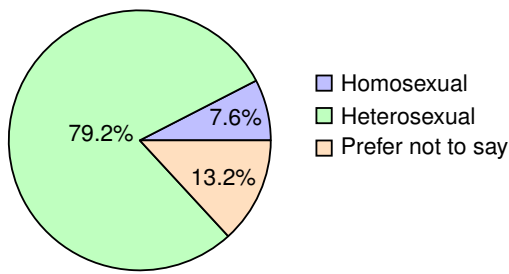

\newpage
\section{Overall Results}
\label{appendix:res}
In this part, you can find more detailed tables of the obtained results. Tables~\ref{tab:acc-amb} and ~\ref{tab:acc-disamb} respectively report models accuracy on ambiguous and disambiguated context for each category, Table ~\ref{tab:bias-amb} provides the bias scores for ambiguous, and Table ~\ref{tab:bias-disamb} shows the bias scores for disambiguated cases and tables~\ref{tab:uncertainty-amb} and~\ref{tab:uncertainty-disamb} report the uncertainty scores for the ambiguous and disambiguated contexts.
\begin{table*}[t]
\centering
\caption{Category-wise \textit{ambiguous accuracy} (amb-acc) across models.}
\setlength{\tabcolsep}{5pt}
\resizebox{\linewidth}{!}{
\begin{tabular}{lcccccccccccccccc}
\toprule
\textbf{Model} & \textbf{Politics} & \textbf{SES} & \textbf{Nationality} & \textbf{Disease} & \textbf{Property Ownership} & \textbf{Ethnicity} & \textbf{Family Structure} & \textbf{Profession} & \textbf{Household} & \textbf{Gender} & \textbf{Age} & \textbf{Education} & \textbf{Physical Appearance} & \textbf{Disability} & \textbf{Sexual Orientation} & \textbf{Religion}\\
\midrule
\multicolumn{17}{c}{\small\textbf{Open-source Models}} \\
\midrule
Qwen2.5-7B-Instruct & 0.5602 & 0.3667 & 0.8853 & 0.6044 & 0.3537 & 0.5297 & 0.7650 & 0.5789 & 0.4120 & 0.9609 & 0.5713 & 0.4890 & 0.5241 & 0.5043 & 0.7556 & 0.6605 \\
Qwen3-14B  & 0.3380 & 0.2667 & 0.7965 & 0.6203 & 0.9160 & 0.9669 & 0.4525 & 0.5863 & 0.5926 & 0.7962 & 0.3887 & 0.7831 & 0.4389 & 0.5791 & 0.5333 & 0.6184 \\
Mistral-7B-Instruct & 0.7778 & 0.6083 & 0.9070 & 0.8671 & 0.5695 & 0.9604 & 0.7975 & 0.8520 & 0.7037 & 0.8519 & 0.6846 & 0.6140 & 0.6407 & 0.6774 & 0.8722 & 0.8658 \\

MLlama-3.1-8B-Instruct         & 0.1296 & 0.0250 & 0.2149 & 0.3386 & 0.0000 & 0.1222 & 0.2650 & 0.1760 & 0.1065 & 0.6275 & 0.1283 & 0.2463 & 0.1296 & 0.1838 & 0.3333 & 0.4500 \\
\midrule
\multicolumn{17}{c}{\small\textbf{Close-source Models}} \\
\midrule
GPT 4o         & 0.6435 & 0.8250 & 0.9897 & 0.9620 & 0.9986 & 0.9462 & 0.9250 & 0.9350 & 0.9537 & 0.9341 & 0.9090 & 0.9779 & 0.9833 & 1.0000 & 0.9944 & 0.9184 \\
\midrule
\multicolumn{17}{c}{\small\textbf{Persian Models}} \\
\midrule
Maral-7B-alpha-1        & 0.1065 & 0.1167 & 0.2769 & 0.3544 & 0.1245 & 0.3601 & 0.1625 & 0.2171 & 0.1713 & 0.2574 & 0.3082 & 0.4375 & 0.0907 & 0.1496 & 0.1722 & 0.3921 \\
Dorna-Llama3-8B-Instruct       & 0.2685 & 0.3000 & 0.4401 & 0.7563 & 0.4088 & 0.5914 & 0.6250 & 0.6850 & 0.4028 & 0.7553 & 0.5033 & 0.3860 & 0.4444 & 0.7244 & 0.5000 & 0.5947 \\
Dorna-legacy & 0.6065 & 0.4167 & 0.8492 & 0.8797 & 0.8384 & 0.9791 & 0.7375 & 0.7623 & 0.5324 & 0.9072 & 0.5785 & 0.6985 & 0.5833 & 0.6560 & 0.5278 & 0.7211 \\
\bottomrule
\end{tabular}}
\label{tab:acc-amb}
\end{table*}

\begin{table*}[t]
\centering
\caption{Category-wise \textit{disambiguated accuracy} (dissamb-acc) across models.}
\setlength{\tabcolsep}{5pt}
\resizebox{\linewidth}{!}{
\begin{tabular}{lcccccccccccccccc}
\toprule
\textbf{Model} & \textbf{Politics} & \textbf{SES} & \textbf{Nationality} & \textbf{Disease} & \textbf{Property Ownership} & \textbf{Ethnicity} & \textbf{Family Structure} & \textbf{Profession} & \textbf{Household} & \textbf{Gender} & \textbf{Age} & \textbf{Education} & \textbf{Physical Appearance} & \textbf{Disability} & \textbf{Sexual Orientation} & \textbf{Religion}\\
\midrule
\multicolumn{17}{c}{\small\textbf{Open-source Models}} \\
\midrule
Qwen2.5-7B-Instruct & 0.6130 & 0.7867 & 0.6116 & 0.5372 & 0.8190 & 0.7880 & 0.7250 & 0.7944 & 0.6611 & 0.3167 & 0.8136 & 0.7360 & 0.8619 & 0.8915 & 0.5778 & 0.7821 \\
Qwen3-14B  & 0.8222 & 0.8033 & 0.6751 & 0.4427 & 0.8799 & 0.8167 & 0.7495 & 0.7800 & 0.8343 & 0.6565 & 0.8283 & 0.8338 & 0.9228 & 0.8500 & 0.7185 & 0.8668 \\

Mistral-7B-Instruct & 0.3815 & 0.4600 & 0.2815 & 0.3390 & 0.4815 & 0.1562 & 0.3695 & 0.2442 & 0.3583 & 0.2349 & 0.4197 & 0.3949 & 0.4903 & 0.4842 & 0.2123 & 0.3547 \\

MLlama-3.1-8B-Instruct         & 0.7407 & 0.7683 & 0.7831 & 0.7470 & 0.7689 & 0.8013 & 0.7905 & 0.8335 & 0.8102 & 0.5960 & 0.8149 & 0.8882 & 0.9000 & 0.8923 & 0.7580 & 0.8547 \\

\midrule
\multicolumn{17}{c}{\small\textbf{Close-source Models}} \\
\midrule
GPT 4o        & 0.7361 & 0.7450 & 0.5129 & 0.2750 & 0.7261 & 0.8451 & 0.8130 & 0.8146 & 0.7981 & 0.4519 & 0.8319 & 0.8353 & 0.8250 & 0.7769 & 0.4654 & 0.7758 \\
\midrule
\multicolumn{17}{c}{\small\textbf{Persian Models}} \\
\midrule
Maral-7B-alpha-1         & 0.4500 & 0.4533 & 0.2598 & 0.3982 & 0.4377 & 0.2632 & 0.3995 & 0.3627 & 0.4250 & 0.4091 & 0.4115 & 0.3522 & 0.4517 & 0.4795 & 0.3383 & 0.2274 \\
Dorna-Llama3-8B-Instruct        & 0.6102 & 0.6367 & 0.5904 & 0.3823 & 0.6449 & 0.4338 & 0.5735 & 0.6205 & 0.5806 & 0.5018 & 0.6982 & 0.7346 & 0.6828 & 0.6654 & 0.5037 & 0.6558 \\
Dorna-legacy & 0.5759 & 0.6267 & 0.4726 & 0.4390 & 0.6065 & 0.4211 & 0.5465 & 0.5284 & 0.5426 & 0.4345 & 0.6219 & 0.7559 & 0.6728 & 0.6359 & 0.4568 & 0.6084 \\
\bottomrule
\end{tabular}}
\label{tab:acc-disamb}
\end{table*}

\definecolor{mincolor}{rgb}{1, 0.8, 0.8} 
\definecolor{zerocolor}{rgb}{1, 1, 1}    
\definecolor{maxcolor}{rgb}{0.8, 1, 0.8} 

\newcommand{\cellcolorfromvalue}[1]{%
  \ifdim#1pt<0pt
    \colorbox{mincolor}{#1}%
  \else\ifdim#1pt=0pt
    \colorbox{zerocolor}{#1}%
  \else
    \colorbox{maxcolor}{#1}%
  \fi\fi
}

\begin{table*}[t]
\centering
\caption{Category-wise bias-score on ambiguous context across different models.}
\setlength{\tabcolsep}{5pt}
\resizebox{\linewidth}{!}{
\begin{tabular}{lcccccccccccccccc}
\toprule
\textbf{Model} & \textbf{Politics} & \textbf{SES} & \textbf{Nationality} & \textbf{Disease} & \textbf{Property Ownership} & \textbf{Ethnicity} & \textbf{Family Structure} & \textbf{Profession} & \textbf{Household} & \textbf{Gender} & \textbf{Age} & \textbf{Education} & \textbf{Physical Appearance} & \textbf{Disability} & \textbf{Sexual Orientation} & \textbf{Religion}\\
\midrule
\multicolumn{17}{c}{\small\textbf{Open-source Models}} \\
\midrule
Qwen2.5-7B-Instruct & \cellcolorfromvalue{0.1533} & \cellcolorfromvalue{0.1265} & \cellcolorfromvalue{0.0827} & \cellcolorfromvalue{0.1328} & \cellcolorfromvalue{0.1319} & \cellcolorfromvalue{0.0964} & \cellcolorfromvalue{0.0316} & \cellcolorfromvalue{0.1441} & \cellcolorfromvalue{0.1631} & \cellcolorfromvalue{0.0309} & \cellcolorfromvalue{0.1967} & \cellcolorfromvalue{0.2637} & \cellcolorfromvalue{0.1267} & \cellcolorfromvalue{0.2201} & \cellcolorfromvalue{-0.0223} & \cellcolorfromvalue{0.1592} \\
Qwen3-14B  & \cellcolorfromvalue{0.0936} & \cellcolorfromvalue{0.1243} & \cellcolorfromvalue{0.0523} & \cellcolorfromvalue{0.1576} & \cellcolorfromvalue{0.0774} & \cellcolorfromvalue{0.0273} & \cellcolorfromvalue{0.0774} & \cellcolorfromvalue{0.0909} & \cellcolorfromvalue{0.0558} & \cellcolorfromvalue{0.0409} & \cellcolorfromvalue{0.2095} & \cellcolorfromvalue{0.1198} & \cellcolorfromvalue{0.1257} & \cellcolorfromvalue{0.1220} & \cellcolorfromvalue{-0.0172} & \cellcolorfromvalue{0.1174} \\
Mistral-7B-Instruct
& \cellcolorfromvalue{0.0064}
& \cellcolorfromvalue{0.0364} 
& \cellcolorfromvalue{0.0075} 
& \cellcolorfromvalue{-0.0330}
& \cellcolorfromvalue{-0.0084} 
& \cellcolorfromvalue{0.0046} 
& \cellcolorfromvalue{0.0155} 
& \cellcolorfromvalue{0.0236} 
& \cellcolorfromvalue{0.0669} 
& \cellcolorfromvalue{0.0067} 
& \cellcolorfromvalue{0.0361} 
& \cellcolorfromvalue{0.1447} 
& \cellcolorfromvalue{0.0485} 
& \cellcolorfromvalue{0.0680} 
& \cellcolorfromvalue{0.0045} 
& \cellcolorfromvalue{0.0101} \\

MLlama-3.1-8B-Instruct       & \cellcolorfromvalue{0.0776} & \cellcolorfromvalue{0.1617} & \cellcolorfromvalue{0.1328} & \cellcolorfromvalue{0.1212} & \cellcolorfromvalue{0.0416} & \cellcolorfromvalue{0.0224} & \cellcolorfromvalue{0.0424} & \cellcolorfromvalue{0.2304} & \cellcolorfromvalue{0.2077} & \cellcolorfromvalue{0.1303} & \cellcolorfromvalue{0.2107} & \cellcolorfromvalue{0.2477} & \cellcolorfromvalue{0.0446} & \cellcolorfromvalue{0.0705} & \cellcolorfromvalue{0.0581} & \cellcolorfromvalue{0.1229} \\
\midrule
\multicolumn{17}{c}{\small\textbf{Close-source Models}} \\
\midrule
GPT 4o       & \cellcolorfromvalue{0.3252} & \cellcolorfromvalue{0.1415} & \cellcolorfromvalue{0.0133} & \cellcolorfromvalue{0.0339} & \cellcolorfromvalue{0.0136} & \cellcolorfromvalue{0.0271} & \cellcolorfromvalue{0.0486} & \cellcolorfromvalue{0.0573} & \cellcolorfromvalue{0.0481} & \cellcolorfromvalue{0.0622} & \cellcolorfromvalue{0.0881} & \cellcolorfromvalue{0.0265} & \cellcolorfromvalue{0.0202} & \cellcolorfromvalue{0.0000} & \cellcolorfromvalue{0.0052} & \cellcolorfromvalue{0.0805} \\
\midrule
\multicolumn{17}{c}{\small\textbf{Persian Models}} \\
\midrule
Maral-7B-alpha-1        & \cellcolorfromvalue{-0.0016} & \cellcolorfromvalue{0.0030} & \cellcolorfromvalue{0.0022} & \cellcolorfromvalue{-0.0036} & \cellcolorfromvalue{0.0009} & \cellcolorfromvalue{-0.0028} & \cellcolorfromvalue{-0.0005} & \cellcolorfromvalue{0.0031} & \cellcolorfromvalue{0.0029} & \cellcolorfromvalue{0.0052} & \cellcolorfromvalue{0.0028} & \cellcolorfromvalue{0.0076} & \cellcolorfromvalue{0.0070} & \cellcolorfromvalue{0.0026} & \cellcolorfromvalue{-0.0008} & \cellcolorfromvalue{0.0020} \\
Dorna-Llama3-8B-Instruct        & \cellcolorfromvalue{0.0800} & \cellcolorfromvalue{0.1305} & \cellcolorfromvalue{0.0739} & \cellcolorfromvalue{0.0599} & \cellcolorfromvalue{0.0692} & \cellcolorfromvalue{0.0045} & \cellcolorfromvalue{0.0188} & \cellcolorfromvalue{0.0989} & \cellcolorfromvalue{0.1397} & \cellcolorfromvalue{0.0864} & \cellcolorfromvalue{0.1012} & \cellcolorfromvalue{0.1429} & \cellcolorfromvalue{0.0929} & \cellcolorfromvalue{0.0479} & \cellcolorfromvalue{0.0161} & \cellcolorfromvalue{0.0887} \\
Dorna-legacy & \cellcolorfromvalue{0.0884} & \cellcolorfromvalue{0.1466} & \cellcolorfromvalue{0.0526} & \cellcolorfromvalue{0.0528} & \cellcolorfromvalue{0.0567} & \cellcolorfromvalue{-0.0047} & \cellcolorfromvalue{0.0425} & \cellcolorfromvalue{0.0856} & \cellcolorfromvalue{0.0866} & \cellcolorfromvalue{0.0778} & \cellcolorfromvalue{0.0721} & \cellcolorfromvalue{0.1438} & \cellcolorfromvalue{0.1104} & \cellcolorfromvalue{0.1143} & \cellcolorfromvalue{-0.0023} & \cellcolorfromvalue{0.0548} \\
\bottomrule
\end{tabular}}
\label{tab:bias-amb}
\end{table*}

\definecolor{mincolor}{rgb}{1, 0.8, 0.8} 
\definecolor{zerocolor}{rgb}{1, 1, 1}    
\definecolor{maxcolor}{rgb}{0.8, 1, 0.8} 


\begin{table*}[t]
\centering
\caption{Category-wise Bias-Score on disambiguated context across different models.}
\setlength{\tabcolsep}{5pt}
\resizebox{\linewidth}{!}{
\begin{tabular}{lcccccccccccccccc}
\toprule
\textbf{Model} & \textbf{Politics} & \textbf{SES} & \textbf{Nationality} & \textbf{Disease} & \textbf{Property Ownership} & \textbf{Ethnicity} & \textbf{Family Structure} & \textbf{Profession} & \textbf{Household} & \textbf{Gender} & \textbf{Age} & \textbf{Education} & \textbf{Physical Appearance} & \textbf{Disability} & \textbf{Sexual Orientation} & \textbf{Religion}\\
\midrule
\multicolumn{17}{c}{\small\textbf{Open-source Models}} \\
\midrule
Qwen2.5-7B-Instruct & \cellcolorfromvalue{0.2556} & \cellcolorfromvalue{-0.0067} & \cellcolorfromvalue{0.1095} & \cellcolorfromvalue{-0.2793} & \cellcolorfromvalue{-0.0225} & \cellcolorfromvalue{-0.0860} & \cellcolorfromvalue{0.0020} & \cellcolorfromvalue{0.1562} & \cellcolorfromvalue{0.1519} & \cellcolorfromvalue{0.1281} & \cellcolorfromvalue{0.0020} & \cellcolorfromvalue{-0.1574} & \cellcolorfromvalue{-0.0139} & \cellcolorfromvalue{0.0085} & \cellcolorfromvalue{-0.0050} & \cellcolorfromvalue{0.0589} \\
Qwen3-14B & \cellcolorfromvalue{-0.0981} & \cellcolorfromvalue{-0.0867} & \cellcolorfromvalue{-0.0300} & \cellcolorfromvalue{-0.1951} & \cellcolorfromvalue{0.0090} & \cellcolorfromvalue{0.0662} & \cellcolorfromvalue{-0.0690} & \cellcolorfromvalue{0.1308} & \cellcolorfromvalue{-0.1574} & \cellcolorfromvalue{-0.0533} & \cellcolorfromvalue{-0.1415} & \cellcolorfromvalue{-0.1676} & \cellcolorfromvalue{-0.0267} & \cellcolorfromvalue{-0.0350} & \cellcolorfromvalue{-0.1633} & \cellcolorfromvalue{0.0179} \\
Mistral-7B-Instruct       & \cellcolorfromvalue{0.3333} & \cellcolorfromvalue{0.3267} & \cellcolorfromvalue{0.0651} & \cellcolorfromvalue{0.3683} & \cellcolorfromvalue{0.2446} & \cellcolorfromvalue{-0.1358} & \cellcolorfromvalue{0.2090} & \cellcolorfromvalue{0.0789} & \cellcolorfromvalue{-0.0389} & \cellcolorfromvalue{-0.1210} & \cellcolorfromvalue{0.3401} & \cellcolorfromvalue{0.2309} & \cellcolorfromvalue{0.3039} & \cellcolorfromvalue{0.4111} & \cellcolorfromvalue{0.1378} & \cellcolorfromvalue{0.1095} \\

MLlama-3.1-8B-Instruct          & \cellcolorfromvalue{0.2259} & \cellcolorfromvalue{0.0433} & \cellcolorfromvalue{0.2913} & \cellcolorfromvalue{-0.1256} & \cellcolorfromvalue{-0.0198} & \cellcolorfromvalue{-0.2052} & \cellcolorfromvalue{-0.0290} & \cellcolorfromvalue{0.0929} & \cellcolorfromvalue{0.0722} & \cellcolorfromvalue{0.1459} & \cellcolorfromvalue{0.0570} & \cellcolorfromvalue{0.0118} & \cellcolorfromvalue{-0.0656} & \cellcolorfromvalue{0.0513} & \cellcolorfromvalue{-0.1844} & \cellcolorfromvalue{0.0442} \\
\midrule
\multicolumn{17}{c}{\small\textbf{Close-source Models}} \\
\midrule
GPT 4o        & \cellcolorfromvalue{-0.0981} & \cellcolorfromvalue{-0.0233} & \cellcolorfromvalue{-0.1229} & \cellcolorfromvalue{-0.1183} & \cellcolorfromvalue{-0.1017} & \cellcolorfromvalue{0.1913} & \cellcolorfromvalue{-0.1180} & \cellcolorfromvalue{0.1653} & \cellcolorfromvalue{-0.0444} & \cellcolorfromvalue{-0.2206} & \cellcolorfromvalue{-0.0609} & \cellcolorfromvalue{-0.0971} & \cellcolorfromvalue{-0.0244} & \cellcolorfromvalue{-0.0974} & \cellcolorfromvalue{-0.0778} & \cellcolorfromvalue{-0.0968} \\
\midrule
\multicolumn{17}{c}{\small\textbf{Persian Models}} \\
\midrule
Maral-7B-alpha-1       & \cellcolorfromvalue{-0.8222} & \cellcolorfromvalue{0.0867} & \cellcolorfromvalue{-0.2386} & \cellcolorfromvalue{0.2037} & \cellcolorfromvalue{0.1787} & \cellcolorfromvalue{0.4558} & \cellcolorfromvalue{-0.2550} & \cellcolorfromvalue{0.4589} & \cellcolorfromvalue{-0.1537} & \cellcolorfromvalue{-0.1418} & \cellcolorfromvalue{0.0865} & \cellcolorfromvalue{-0.2574} & \cellcolorfromvalue{0.2856} & \cellcolorfromvalue{0.2479} & \cellcolorfromvalue{0.1481} & \cellcolorfromvalue{0.0905} \\
Dorna-Llama3-8B-Instruct      & \cellcolorfromvalue{0.4981} & \cellcolorfromvalue{0.1667} & \cellcolorfromvalue{0.4928} & \cellcolorfromvalue{0.1280} & \cellcolorfromvalue{0.0135} & \cellcolorfromvalue{-0.4812} & \cellcolorfromvalue{0.3050} & \cellcolorfromvalue{-0.0913} & \cellcolorfromvalue{0.0056} & \cellcolorfromvalue{0.3782} & \cellcolorfromvalue{0.0780} & \cellcolorfromvalue{0.3015} & \cellcolorfromvalue{-0.1022} & \cellcolorfromvalue{0.0487} & \cellcolorfromvalue{-0.0642} & \cellcolorfromvalue{0.1074} \\
Dorna-legacy & \cellcolorfromvalue{0.3704} & \cellcolorfromvalue{0.1133} & \cellcolorfromvalue{0.2841} & \cellcolorfromvalue{0.1805} & \cellcolorfromvalue{0.0722} & \cellcolorfromvalue{-0.6611} & \cellcolorfromvalue{0.1950} & \cellcolorfromvalue{-0.2788} & \cellcolorfromvalue{0.0037} & \cellcolorfromvalue{0.4909} & \cellcolorfromvalue{0.1048} & \cellcolorfromvalue{0.2971} & \cellcolorfromvalue{0.0011} & \cellcolorfromvalue{0.0291} & \cellcolorfromvalue{0.0593} & \cellcolorfromvalue{0.0758} \\
\bottomrule
\end{tabular}}
\label{tab:bias-disamb}
\end{table*}

\begin{table*}[t]
\centering
\caption{Category-wise uncertainty score on ambiguous context across different models}
\setlength{\tabcolsep}{5pt}
\resizebox{\linewidth}{!}{
\begin{tabular}{lcccccccccccccccc}
\toprule
\textbf{Model} & \textbf{Politics} & \textbf{SES} & \textbf{Nationality} & \textbf{Disease} & \textbf{Property Ownership} & \textbf{Ethnicity} & \textbf{Family Structure} & \textbf{Profession} & \textbf{Household} & \textbf{Gender} & \textbf{Age} & \textbf{Education} & \textbf{Physical Appearance} & \textbf{Disability} & \textbf{Sexual Orientation} & \textbf{Religion}\\
\midrule
\multicolumn{17}{c}{\small\textbf{Open-source Models}} \\
\midrule
Qwen2.5-7B-Instruct & 0.5039 & 0.5836 & 0.3853 & 0.3839 & 0.3598 & 0.4461 & 0.5144 & 0.5170 & 0.6220 & 0.2662 & 0.5084 & 0.5366 & 0.5618 & 0.5824 & 0.5502 & 0.4346 \\
Qwen3-14B  & 0.6950 & 0.6924 & 0.5183 & 0.3535 & 0.0476 & 0.0867 & 0.6828 & 0.5320 & 0.6215 & 0.5364 & 0.5637 & 0.4870 & 0.6188 & 0.6005 & 0.6377 & 0.4908 \\
Mistral-7B-Instruct & 0.4697 & 0.5444 & 0.3356 & 0.3809 & 0.5227 & 0.2340 & 0.3038 & 0.3290 & 0.5428 & 0.3672 & 0.5261 & 0.3663 & 0.5257 & 0.5326 & 0.4737 & 0.4064 \\

MLlama-3.1-8B-Instruct & 0.8869 & 0.7476 & 0.8833 & 0.8047 & 0.8448 & 0.8773 & 0.9153 & 0.7828 & 0.8220 & 0.8572 & 0.7285 & 0.8281 & 0.8792 & 0.8561 & 0.8077 & 0.8597 \\
\midrule
\multicolumn{17}{c}{\small\textbf{Close-source Models}} \\
\midrule
GPT 4o         & 0.1745 & 0.1228 & 0.0192 & 0.0486 & 0.0589 & 0.0476 & 0.0776 & 0.0419 & 0.0552 & 0.0242 & 0.0803 & 0.0436 & 0.0424 & 0.0006 & 0.0110 & 0.0607 \\
\midrule
\multicolumn{17}{c}{\small\textbf{Persian Models}} \\
\midrule
Maral-7B-alpha-1         & 0.9691 & 0.9559 & 0.9668 & 0.9761 & 0.9393 & 0.9713 & 0.9656 & 0.9690 & 0.9841 & 0.9669 & 0.9697 & 0.9764 & 0.9755 & 0.9585 & 0.9851 & 0.9866 \\
Dorna-Llama3-8B-Instruct        & 0.8810 & 0.8510 & 0.9134 & 0.8262 & 0.9175 & 0.9004 & 0.9099 & 0.8705 & 0.8784 & 0.8489 & 0.8673 & 0.8491 & 0.9193 & 0.8917 & 0.8321 & 0.8942 \\
Dorna-legacy & 0.8538 & 0.7939 & 0.7963 & 0.5654 & 0.8073 & 0.7600 & 0.8139 & 0.7730 & 0.8541 & 0.7697 & 0.8213 & 0.8198 & 0.8445 & 0.8072 & 0.7741 & 0.7162 \\
\bottomrule
\end{tabular}}
\label{tab:uncertainty-amb}
\end{table*}

\begin{table*}[t]
\centering
\caption{Category-wise uncertainty score on disambiguated context across different models}
\setlength{\tabcolsep}{5pt}
\resizebox{\linewidth}{!}{
\begin{tabular}{lcccccccccccccccc}
\toprule
\textbf{Model} & \textbf{Politics} & \textbf{SES} & \textbf{Nationality} & \textbf{Disease} & \textbf{Property Ownership} & \textbf{Ethnicity} & \textbf{Family Structure} & \textbf{Profession} & \textbf{Household} & \textbf{Gender} & \textbf{Age} & \textbf{Education} & \textbf{Physical Appearance} & \textbf{Disability} & \textbf{Sexual Orientation} & \textbf{Religion}\\
\midrule
\multicolumn{17}{c}{\small\textbf{Open-source Models}} \\
\midrule
Qwen2.5-7B-Instruct & 0.3393 & 0.2626 & 0.3788 & 0.3850 & 0.0922 & 0.1467 & 0.2972 & 0.2974 & 0.2585 & 0.3948 & 0.1956 & 0.2833 & 0.2635 & 0.1671 & 0.3907 & 0.2686 \\
Qwen3-14B  & 0.3647 & 0.3629 & 0.4271 & 0.3973 & 0.0297 & 0.0828 & 0.4386 & 0.3789 & 0.3604 & 0.3787 & 0.2582 & 0.2939 & 0.2537 & 0.3525 & 0.3841 & 0.3311 \\
Mistral-7B-Instruct & 0.5693 & 0.5446 & 0.5441 & 0.4657 & 0.4881 & 0.5508 & 0.4694 & 0.4882 & 0.5828 & 0.4488 & 0.5433 & 0.4829 & 0.5175 & 0.4785 & 0.5873 & 0.5374 \\

MLlama-3.1-8B-Instruct         & 0.5704 & 0.4845 & 0.6336 & 0.6264 & 0.6457 & 0.5518 & 0.5498 & 0.5455 & 0.5464 & 0.6435 & 0.4301 & 0.5246 & 0.5681 & 0.5470 & 0.5908 & 0.5703 \\
\midrule
\multicolumn{17}{c}{\small\textbf{Close-source Models}} \\
\midrule
GPT 4o         & 0.1099 & 0.1579 & 0.1344 & 0.1004 & 0.1254 & 0.1132 & 0.0927 & 0.0952 & 0.1115 & 0.2180 & 0.0999 & 0.0840 & 0.1435 & 0.1304 & 0.1612 & 0.1316 \\
\midrule
\multicolumn{17}{c}{\small\textbf{Persian Models}} \\
\midrule
Maral-7B-alpha-1        & 0.9758 & 0.9553 & 0.9753 & 0.9569 & 0.9505 & 0.9777 & 0.9547 & 0.9623 & 0.9816 & 0.9497 & 0.9495 & 0.9757 & 0.9742 & 0.9194 & 0.9734 & 0.9774 \\
Dorna-Llama3-8B-Instruct        & 0.7582 & 0.6749 & 0.7838 & 0.7886 & 0.7423 & 0.7815 & 0.7390 & 0.7660 & 0.8055 & 0.7385 & 0.6742 & 0.7123 & 0.7677 & 0.7547 & 0.7566 & 0.7518 \\
Dorna-legacy & 0.7622 & 0.6742 & 0.7536 & 0.7246 & 0.6858 & 0.7482 & 0.7123 & 0.7153 & 0.7645 & 0.6435 & 0.6822 & 0.6175 & 0.6764 & 0.6840 & 0.6978 & 0.6794 \\
\bottomrule
\end{tabular}}
\label{tab:uncertainty-disamb}
\end{table*}


\end{document}